\pdfoutput=1

\documentclass[11pt]{article}

\usepackage[preprint]{acl}

\usepackage{times}
\usepackage{latexsym}
\usepackage{amsmath}

\usepackage[T1]{fontenc}

\usepackage[utf8]{inputenc}

\usepackage{microtype}

\usepackage{inconsolata}

\usepackage{graphicx}
\usepackage{tabularx}
\newcolumntype{M}[1]{>{\centering\arraybackslash}m{#1}} 
\usepackage{multirow}
\usepackage{booktabs}       
\usepackage{amsfonts}       
\usepackage{nicefrac}       
\usepackage{microtype}      
\usepackage{xcolor}         
\usepackage{array}          
\usepackage{longtable}      
\usepackage{makecell}       
\usepackage{graphicx}       
\usepackage{subcaption}     
\usepackage{lipsum}
\usepackage{xtab,afterpage}
\usepackage{balance}
\usepackage{tikzpagenodes}
\usepackage{hhline}

\usepackage[T1]{fontenc}

\usepackage[utf8]{inputenc}

\usepackage{microtype}
\newcolumntype{M}[1]{>{\centering\arraybackslash}m{#1}} 

\title{TTQA-RS- A break-down prompting approach for Multi-hop Table-Text Question Answering with Reasoning and Summarization}

 \author{Jayetri Bardhan, Bushi Xiao, Daisy Zhe Wang \\
        Department of Computer and Information Science and Engineering \\ University of Florida \\ Gainesville, Florida, USA \\
       \{jayetri.bardhan, xiaobushi, daisyw\}@ufl.edu  } 
       
\begin{document}
\maketitle
\begin{abstract}
Question answering (QA) over tables and text has gained much popularity over the years. Multi-hop table-text QA requires multiple hops between the table and text, making it a challenging QA task. Although several works have attempted to solve the table-text QA task, most involve training the models and requiring labeled data. In this paper, we have proposed a Retrieval Augmented Generation (RAG) based model - “TTQA-RS: A break-down prompting approach for Multi-hop Table-Text Question Answering with Reasoning and Summarization”\footnote{Link to code in Github: \url{https://github.com/jayetri/TTQA-RS}}. Our model uses an enhanced retriever for table-text information retrieval and uses augmented knowledge, including table-text summary with decomposed sub-questions with answers for a reasoning-based table-text QA. Using open-source language models, our model outperformed all existing prompting methods for table-text QA tasks on existing table-text QA datasets, such as HybridQA and OTT-QA's development set. Our experiments demonstrate the potential of prompt-based approaches using open-source LLMs. Additionally, by using LLaMA3-70B, our model achieved state-of-the-art performance for prompting-based methods on multi-hop table-text QA.

\end{abstract}

\section{Introduction}

Question Answering over tables involves extracting the table cell containing the answer to the question. The most popular approach of table QA is to generate SQL queries using the question, i.e. the table-QA task is converted into a text-to-SQL task \citep{pasupat-liang-2015-compositional, yu-etal-2018-spider, zhong2017seq2sql}. The SQL queries are then used to retrieve the answer from the tables. Some other recent approaches use an intermediate pre-training method on the flattened tables for QA \citep{herzig-etal-2020-tapas, yin-etal-2020-tabert}. QA over table and text is more challenging. Datasets like HybridQA \citep{chen2020hybridqa} and OTT-QA \citep{chen2020open} are examples of multi-hop table-text QA datasets where the answer to the question can exist in the table or the text. These two datasets make use of Wikitables along with text from Wikipedia to answer the questions.  The tables in the HybridQA dataset contain hyperlinks linking the table cells to Wikipedia’s text, making QA tasks more challenging. Additionally, HybridQA and OTT-QA are both multi-hop table-text datasets, which means that one or more hops between the table and text are required to derive the answer. 

\begin{figure*}[hbt!]
    \centering
    \includegraphics[scale=0.9]{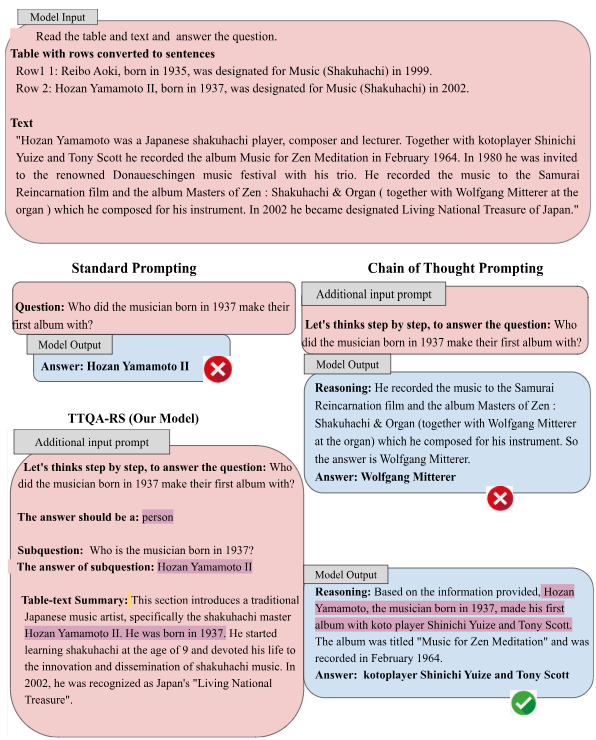}
    \caption{Comparison between Standard prompting, Chain of Thought prompting, and the TTQA-RS model.}
    \vspace{-1.5em}
    \label{fig:Comparison between Standard prompting, Chain of Thought prompting, and the TTQA-RS model}
\end{figure*}

Over the years, several works have attempted to solve this task. But the majority of these works have used supervised-training for the reader, requiring a large amount of labeled data \citep{chen2020hybridqa, sun2021end, wang-etal-2022-muger2, eisenschlos-etal-2021-mate, feng-etal-2022-multi, kumar-etal-2023-multi, chen2020open, li-etal-2021-dual}. In this paper, we have proposed a prompting-based approach while using open-source large language models (LLMs) for multi-hop table-text QA. 

With the emergence of new generative-based LLM models, prompt-based methods using in-context learning have started being explored \citep{chen2023large}. Training models from scratch or even fine-tuning the models requires a large amount of labeled data. In-context learning is a cheaper alternative approach that does not need any fine-tuning but instead uses pre-trained language models (LLMs) to solve new tasks using a few examples as part of the prompt. The release of the new openAI models such as GPT 4 has opened new avenues of research in natural language processing and has encouraged further research in prompt learning. \citep{wei2022chain} has shown that reasoning with chain of thought (CoT) can significantly improve the ability of large language models to perform complex reasoning in tasks including QA. But small LLMs, i.e. models with less than 100B parameters using CoT prompting tend to hallucinate and produce incorrect results, urging research communities to use bigger LLMs which are expensive and also not open-source.

 In this paper, we introduced a framework - TTQA-RS, a prompting approach for table-text QA that despite CoT’s shortcomings on small-parameter models, we were able to reduce the hallucinations on open-source small models. Furthermore, our proposed model was able to beat the state-of-the-art model - S3HQA’s CoT prompting using different LLaMA models \citep{lei-etal-2023-s3hqa} on HybridQA dataset. By beating their model’s performance, we have shown the potential for smaller LLMs in multi-hop table-text QA.

For our experiments, we have used HybridQA dataset and OTT-QA's development set. OTT-QA is an extension of the HybridQA dataset. Similar to the HybridQA dataset, the OTT-QA dataset is also constructed using questions based on Wikipedia tables and text. But unlike the HybridQA dataset, the test set of the OTT-QA dataset does not have hyperlinks in the table cells that can be linked to the Wikipedia text. Hence, the OTT-QA's test set is more challenging. The task of linking tables and text passages for open-domain QA is out of scope of this paper. 

The TTQA-RS model breaks down the table-text QA problem into multiple steps. In the HybridQA and the OTT-QA  dataset, the questions require multiple steps of reasoning over table and text to answer. The TTQA-RS model generates the sub-questions that can help in answering the complex questions. It also generates the summary of the table and text, which is in turn used for the table-text QA of the original questions. Breaking down the complex multi-hop QA problem into simple, smaller steps along with providing augmented information, including the table-text summary, can improve the performance of multi-hop table-text QA tasks using small open-source LLMs. We have also proposed an enhanced retriever for our framework that further improves the overall performance of our table-text QA model. In Figure~\ref{fig:Comparison between Standard prompting, Chain of Thought prompting, and the TTQA-RS model}, we show an example of a question that uses standard prompting, CoT, and the TTQA-RS approach for multi-hop QA.

\section{Related Works}

Multi-hop table-text QA can be a complex task as it requires multiple hops between the table and text to answer the questions. S3HQA \citep{lei-etal-2023-s3hqa} and MFORT-QA \citep{guan2024mfort} are the only two existing models as per our knowledge that use in-context learning for multi-hop table-text QA. The S3HQA model has demonstrated table-text QA task using the Hybrid-QA dataset, whereas MFORT-QA has used the OTT-QA dataset. The S3HQA model uses a three-step method - a retriever with refinement training, a hybrid selector, and a generation-based reasoner with GPT 3.5 for the hybrid table-text QA task. MFORT-QA uses the Chain-of-thought (CoT) method to break down complex questions into smaller sub-questions, and uses Retrieval Augmented Generation to extract more context. Similar to the MFORT-QA model, we also use a RAG-based framework and break down complex questions into smaller sub-questions. With the complexity of the multi-hop QA task broken down into smaller questions, LLMs are in turn working on a smaller problem and perform better as single-step reasoners. Our model - TTQA-RS, additionally uses an enhanced retriever for table-text information retrieval and further generates a summary using the retrieved table rows and passages. Then, for table-text question answering (QA), it uses the generated summary, the predicted entity type of the answer, and the generated sub-questions along with the answer.

\section{Our Model}
\subsection{System Overview}

\begin{figure*}
    \centering
 \includegraphics[scale=0.45]{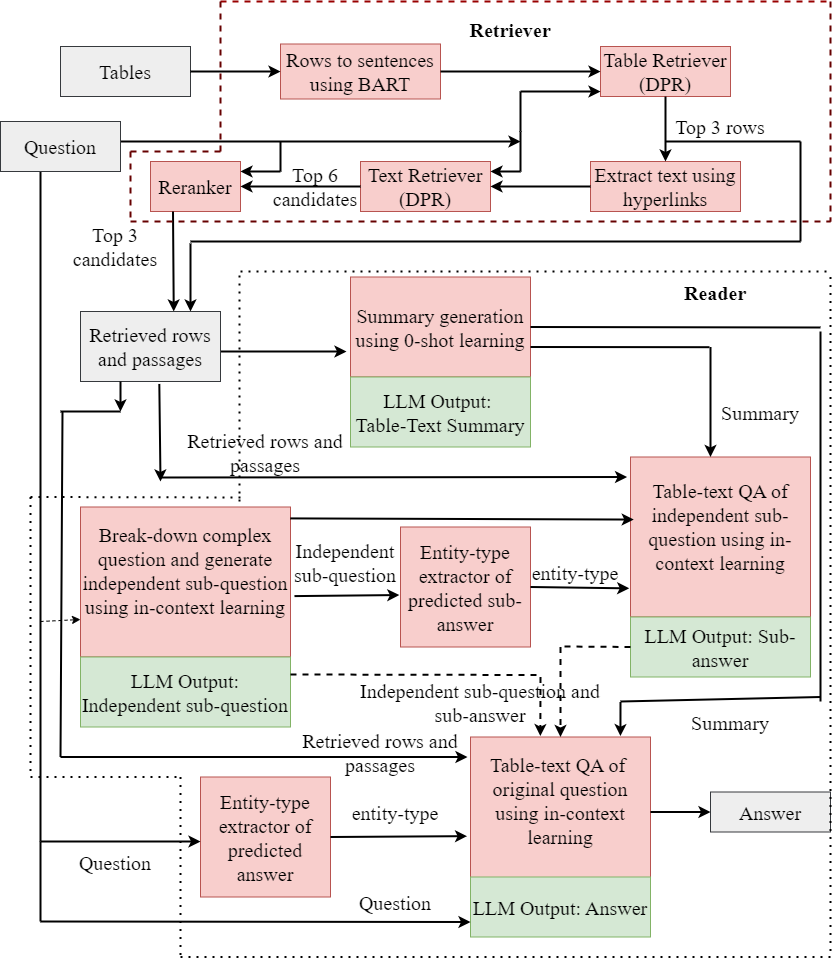}
    \caption{An overview of TTQA-RS framework. The brown dashed lines represent the retriever and the black dotted lines represent the reader for the table-text QA model.}
     \vspace{-1em}
    \label{fig:An overview of TTQA-RS framework}
\end{figure*}

The TTQA-RS model uses a RAG-based retriever-reader framework. Our reader breaks down the table-text QA problem into five steps - (1) Summary generation using retrieved tables rows and passages, (2) Question decomposition, (3) Entity type prediction of the expected answer, (4) Table-text QA of independent sub-question, and (5) Table-text QA of the original question.  Figure~\ref{fig:An overview of TTQA-RS framework} shows an overview of the TTRS-QA framework. 

\subsection{Retriever}

\begin{figure*}
    \centering
    \includegraphics[scale=0.75]{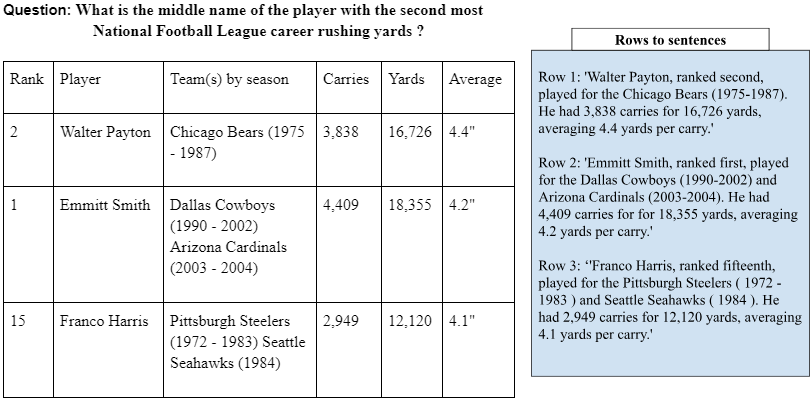}
    \caption{Example of rows to sentences for a table}
    \label{fig:Example of converting rows to sentences using Large BART}
     \vspace{-2em}
\end{figure*}

We have used an enhanced retriever that first converts rows in the tables to sentences using a pre-trained BART-Large model. The questions and sentences are encoded and then fed to a fine-tuned Dense Passage Retriever (DPR) (we can call this the table retriever). The table retriever retrievers the top 3 most relevant rows. It is important to first convert the rows to sentences using BART since the retriever is not always smart enough to interpret the rows for retrieval. This step helps the retriever to make more intelligent decisions. Figure \ref{fig:Example of converting rows to sentences using Large BART} shows an example where the rows of the table are converted to sentences. In this example, when the table is directly fed to the retriever, it fails to understand that rank 2 implies that the player came second in the National Football League. When the rows are converted to sentences before feeding to the DPR, it can better understand the table information and, in turn, retrieve more accurate information.

For text retrieval, we extract the passages from the text linked to the top 3 rows using hyperlinks. The question embedding and row embedding refer to the final dense vectors generated from the encoding process of the question and the rows, respectively. We calculate the combined embeddings using the following: 

\begin{align}
\label{eq:combined_embeddings}
\text{combined\_embedding} &= \alpha \cdot \text{question\_embedding} \nonumber \\
&+ (1 - \alpha) \cdot \text{row\_embedding}
\end{align}

\begin{figure*}[h]
    \centering
    \includegraphics[scale=0.6]{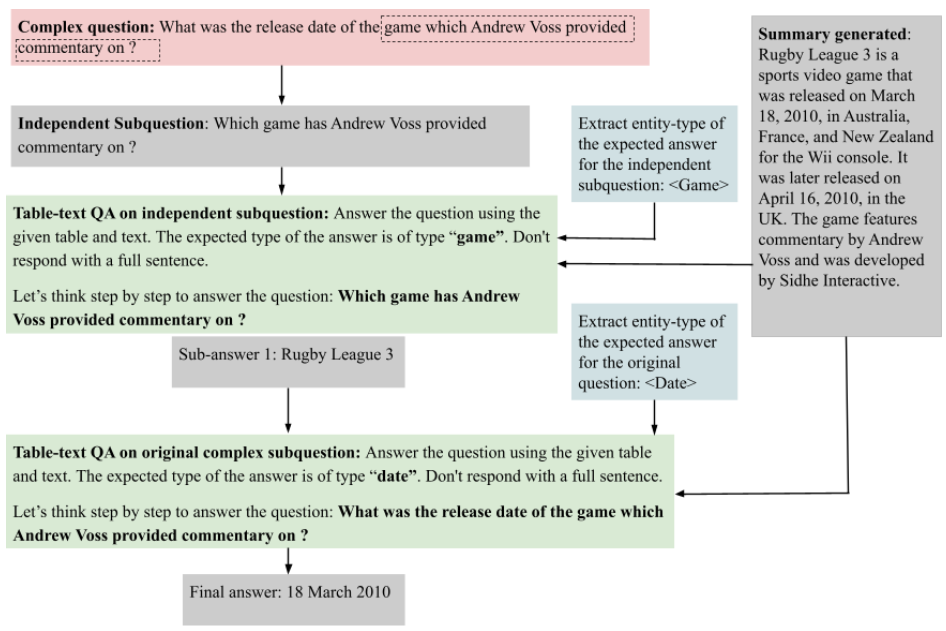}
    \caption{Example of our approach using TTQA-RS model}
    \label{fig:Example of our approach using TTQA-RS model}
     \vspace{-1.5em}
\end{figure*}

In the Equation~\ref{eq:combined_embeddings}, $\alpha$ is a parameter with values between 0 and 1. We estimate the similarity between the combined embeddings and the text embeddings to retrieve the top 6 candidates using a pre-trained DPR model (i.e. the text retriever). These top 6 candidates, along with the question keywords generated using a pre-trained T5 model, are fed into a re-ranker, (a pre-trained 'deepset/roberta-base-squad2' model). The re-ranker identifies the top 3 candidates, which are subsequently forwarded to the reader.

\subsection{Reader}

\subsubsection{Table-text Summarization}
This is the first step of the reader model. The retrieved rows and passages are used to generate summaries of the table and text. We used zero-shot learning with LLaMA 3-70B model to generate the summaries. In Appendix \ref{Table-text summarization} we have shown an example of a table-text summarization prompt.

\subsubsection{Question decomposition}
In the next step, we break down the questions and identify the sub-questions, such that the answer of one sub-question can aid in answering the original complex question. From here onwards, we will refer to the sub-question that can be answered first as the “independent sub-question”. Let's take the example in Figure~\ref{fig:Example of our approach using TTQA-RS model}. The complex question - "What was the release date of the game which Andrew Voss provided commentary on ?" can be broken down into sub-questions. The independent sub-question for this question is - "Which game has Andrew Voss provided commentary on?". The answer to this sub-question is "Rugby League 3". This can be used to simplify the original complex question to the following - " What was the release date of Rugby League 3?". Thus, including the information about the independent sub-question and the sub-answer helps to reduce the complexity of the multi-hop task. Identifying the independent sub-question and breaking down the complex multi-hop QA problem helps to reduce the complexity of the problem, and in turn, boosts the accuracy of the model. We use in-context learning with LLaMA3-70B model to generate the independent sub-questions for the given complex queries. Appendix~\ref{Independent sub-questions} shows further examples.

\subsubsection{Entity type prediction of the expected answer}
We identify the entity type of the expected answer for both the independent sub-question and also for the original question. For the following question - "What was the release date of the game which Andrew Voss provided commentary on?", the entity type of the expected answer is "date". Knowing that the expected answer is of type - "date", makes the LLM's task of generating the answer considerably easier. We have used Spacy, an open-source Python library to obtain the entity type.   

\subsubsection{Table-text QA of independent sub-questions}
In this step, we use few-shot learning with  CoT to generate the answers of the independent sub-questions. The input prompt contains the retrieved table rows, retrieved passages, the table-text summary, and also the predicted entity type of the expected answer. This is used to generate the answer for the independent sub-question.

\subsubsection{Table-text QA of the original questions}
This is the final step of the table-text QA framework. To generate the answers of the original questions, we use CoT-based in-context learning similar to the previous step. But in addition to the prompt containing the retrieved rows, retrieved passages, table-text summary, and the expected entity type of the predicted answer of the original question, it also includes the independent sub-question with its generated sub-answer obtained in the previous step. Figure~\ref{fig:Example of our approach using TTQA-RS model} shows an example of our reader's approach. For simplicity, we have excluded mentioning about the few-shot examples in Figure~\ref{fig:Example of our approach using TTQA-RS model}.

\section{Experimental Setup}

\subsection{Datasets}

\textbf{HybridQA}
HybridQA \citep{chen2020hybridqa} is a large QA dataset that requires multi-hop reasoning over tables and text for QA. The questions in the HybridQA dataset are based on Wikipedia tables and corpora that are linked to the Wikipedia tables through hyperlinks.

\textbf{OTT-QA} \citep{chen2020open} is an open-domain multi-hop table-text QA dataset. For our experiments, we only use the development set of the OTT-QA dataset.

\subsection{Implementation details}

The implementation details are shown in Appendix~\ref{Implementation Details}.

\subsection{Baseline Models}
\textbf{Standard prompting} - For the baseline standard prompting model, we used the same retriever as in TTQA-RS model. For the reader, we performed in-context learning with standard prompting \citep{brown2020language} for the QA task.

\textbf{Chain of Thought Prompting (CoT)} - The CoT baseline model also uses the same retriever as the TTQA-RS model. The reader uses in-context learning with CoT prompting \cite{wei2022chain}. 

\textbf{Least-to-Most Prompting} - Least-to-Most Prompting (LtM) \cite{zhou2022least} extends CoT prompting by first diving a problem into subproblems and then solving each one. For the baseline LtM model, we use TTQA-RS's retriever with LtM prompting \cite{zhou2022least}.

\textbf{CoT with Self Consistency (sc)} - \cite{wang2022self} proposed self-consistency as a replacement for the naive greedy decoding utilized in CoT prompting.

\section{Results and Discussion}

\subsection{Main Results}

\begin{table}
  \centering
    \caption{\label{Retrieval Accuracy} Retrieval Accuracy (\%)}
  \begin{tabular}{M{1.2cm}| M{1.5cm}| M{1.5cm}| M{1.7cm}}
    \hline
    \multirow{2}{*}{\bfseries Metrics}  &  \multicolumn{2}{c|}{\bfseries HybridQA} &  \multicolumn{1}{c}{\bfseries OTT-QA} \\
            \cline{2-4}

     &   \multicolumn{1}{c|}{Dev}  & \multicolumn{1}{c|}{ Test}  &  \multicolumn{1}{c}{ Dev}  \\
    \hline
   HIT@1    &  73.28 & 71.49 & 72.53 \\
   \hline
   HIT@3 &   82.01 & 80.65 & 81.83\\
    \hline
  \end{tabular}
     \vspace{-2em}

\end{table}

\begin{table*}[hbt!]
  \centering
    \caption{Performance of our model-TTQA-RS and other related works on the HybridQA dataset}
\label{tab:Performance of our model-TTQA-RS and other related works on the HybridQA dataset}
  \begin{tabular}{M{1cm}|M{8cm}| M{2cm} | M{2cm}}
    \hline
    \textbf{Type} & \textbf{Model}          & \textbf{Dev} & \textbf{Test} \\
                   &   &  \textbf{EM / F1 } & \textbf{EM / F1 } \\
    \hline
    Train & HYBRIDER \citep{chen2020hybridqa}       &  44.0 / 50.7    &   43.8 / 50.6   \\
    Train & DocHopper \citep{sun2021end} & 47.7 / 55.0 &  46.3 / 53.3 \\
    Train & MuGER2 \citep{wang-etal-2022-muger2} & 57.11 / 67.3 & 56.3 / 66.2 \\
    Train & POINTR + MATE \citep{eisenschlos-etal-2021-mate}  & 63.4 / 71.0 & 62.8 / 70.2 \\
    Train & DEHG \citep{feng-etal-2022-multi} & 65.2 / 76.3 & 63.9 / 75.5 \\
    Train & MITQA \citep{kumar-etal-2023-multi} & 65.5 / 72.7 & 64.3 / 71.9 \\
    Train & MAFiD \citep{lee-etal-2023-mafid} & 66.2 / 74.1 & 65.4 / 73.6 \\
    Train & S3HQA (supervised learning) \citep{lei-etal-2023-s3hqa} & 68.4 / 75.3 & 67.9 / 75.5 \\
    2-shot & S3HQA LLaMA 3 - 70B CoT \citep{lei-etal-2023-s3hqa} &  52.23 / 58.49  & 53.75 / 62.94 \\ 
    \hline
    2-shot & Baseline Standard prompting LLaMA 3-70B &   48.73 / 60.02   &   50.76 / 59.23    \\
    2-shot & Baseline CoT LLaMA 3-70B &   54.53 / 62.49    &   55.89 / 62.98   \\
    2-shot & Baseline CoT LLaMA 3-70B (with sc) &   54.63 / 64.07    &   55.96 / 63.07  \\
    2-shot & Baseline Least-to-Most prompting LLaMA 3-70B &   57.84 / 65.42    &   57.83 / 67.39   \\
    \hhline{=|=|=|=}
    2-shot & TTQA-RS LLaMA 3 - 70B &  63.97 / 74.20   &  62.34 / 70.58  \\
    2-shot & TTQA-RS LLaMA 3 - 70B (with sc)  &  \textbf{63.98 / 74.32} &  \textbf{62.92 / 70.83}\\ \hline
            & Human &  - & 88.2 / 93.5 \\
    
    \hline
  \end{tabular}

\end{table*}

%


\begin{figure*}
    \centering
    \includegraphics[scale=0.7]{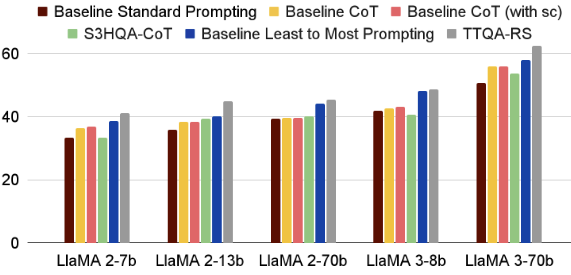}
    \caption{EM-score of HybridQA test set on different LLaMA models}
    \label{fig:5}
     \vspace{-0.5em}
\end{figure*}

\begin{table*}[hbt!]
  \centering
    \caption{\label{Performance of OTT-QA and related works}
Performance of our model - TTQA-RS and other related works on OTT-QA development set.
  }
  \begin{tabular}{M{1cm}|M{9.5cm}| M{2cm} | M{2cm}}
    \hline
    \multirow{2}{*}{\bfseries Type}  & \multirow{2}{*}{\bfseries  Model}    & \multicolumn{2}{c}{\bfseries Dev} \\

    \cline{3-4}
    &   &   \multicolumn{1}{c|}{\bfseries EM } & \multicolumn{1}{c}{\bfseries F1} \\
    \hline
    Train & HYBRIDER (Top-1) \citep{chen2020hybridqa} \citep{chen2020hybridqa} &  8.9   &   11.3   \\
    Train & HYBRIDER (best Top-K)   & 10.3 & 13.0 \\
    Train & Iterative-Retrieval + Single-Block Reader  \citep{chen2020open}   & 7.9 & 11.1 \\
    Train & Fusion-Retrieval + Single-Block Reader \citep{chen2020open} &  13.8  & 17.2 \\
    Train & Iterative-Retrieval + Cross-Block Reader  \citep{chen2020open} & 14.4 & 18.5 \\
    Train & Fusion-Retrieval + Cross-Block Reader \citep{chen2020open} & 28.1 & 32.5 \\
    Train & CARP \citep{zhong2022reasoning}  &  33.2  &  38.6 \\
    Train & MITQA \citep{kumar-etal-2023-multi} & 40.0  &  45.1 \\
    \hline
    2-shot & Baseline Standard prompting LLaMA3-70B & 51.24 &  58.73\\
    2-shot & Baseline CoT LLaMA3-70B & 55.78 & 64.30 \\
    2-shot & Baseline CoT LLaMA3-70B (with sc) & 55.79 & 65.54 \\
    2-shot & Baseline Least-to-Most prompting LLaMA3-70B & 58.68 & 67.75 \\

     \hhline{=|=|=|=}
    2-shot & TTQA-RS LLaMA 3 - 70B &  63.17 & 73.52  \\
    2-shot & TTQA-RS LLaMA 3 - 70B (with sc)  & \textbf{63.31} & \textbf{73.54} \\ 
    
    \hline
  \end{tabular}
     \vspace{-0.5em}

\end{table*}

\begin{table*}
  \centering
    \caption{\label{Ablation studies of TTQA-RS on HybridQA dataset}
Ablation studies of TTQA-RS on HybridQA and OTT-QA dataset using LLaMA 3-70B
  }
  \begin{tabular}{M{9cm}|M{1.9cm}|M{1.9cm}|M{1.9cm}}
    \hline

    \multirow{2}{*}{\bfseries Model}  &  \multicolumn{2}{c|}{\bfseries HybridQA} &  \multicolumn{1}{c}{\bfseries OTT-QA} \\
            \cline{2-4}

     &   \multicolumn{1}{c|}{Dev}  & \multicolumn{1}{c|}{ Test}  &  \multicolumn{1}{c}{ Dev}  \\

    \hline
   Baseline Standard prompting  & 48.73/60.02    &   50.76/59.23   &  51.24/58.73     \\
   \hline
   Baseline CoT & 54.53/62.49 &  55.89/62.98  &  55.78/64.30  \\
   \hline
    CoT + entity-type prediction of expected answer & 55.34/63.95 & 56.52/63.47 &  57.60/67.84 \\
    \hline
   Question decomposition and including sub-question with generated sub-answer + entity-type prediction of expected answer  with COT (no summarization)
 & 60.85/70.68 & 58.32/66.82 & 61.06/70.48  \\
 \hline
   Our model - Question decomposition and including sub-question with generated sub-answer + entity-type prediction of expected answer + summarization with CoT   & 63.97/74.20 & 62.34/70.58 & 63.17/73.52 \\
    
    \hline
  \end{tabular}
     \vspace{-1.5em}

\end{table*}

In this section, we discuss all our major findings. Table~\ref{Retrieval Accuracy} shows the hit score of our model on the HybridQA and OTT-QA dataset. Table~\ref{tab:Performance of our model-TTQA-RS and other related works on the HybridQA dataset} displays the performance of our model with other existing models on the HybridQA dataset.  From Table~\ref{tab:Performance of our model-TTQA-RS and other related works on the HybridQA dataset}, we can observe that most existing models train their models for table-text QA. S3HQA \citep{lei-etal-2023-s3hqa} is the only model among the existing works that uses in-context learning for the HybridQA dataset. Our TTQA-RS model with LLaMA 3-70B on the HybridQA's development set was able to beat S3HQA's CoT model by 11.7\% exact match (EM) and by 15.71\% F1 score. Our model outperformed the S3HQA-CoT model on the HybridQA test using LLaMA3-70B by 8.6\% and 7.6\% exact match (EM) and F1 score respectively. Our 2-shot model with LLaMA 3-70B also beats the baseline standard, baseline CoT,  baseline COT with sc, and Least-to-Most prompting method by a huge margin. This shows the effectiveness of our reader model. We can also observe that our baseline CoT model outperformed the S3HQA's CoT model for the HybridQA dataset. This shows the effectiveness of our retriever model.  Furthermore, in Figure~\ref{fig:5}, we have shown the performance of our TTQA-RS model on different parameter models of LLaMA 2 and LLaMA 3 models on HybridQA test set. For all the different parameter models of LLaMA-2 and LLaMA-3, our framework performed better than the baseline prompting models and also the S3HQA-CoT model. Our experiments show that our approach can improve the performance of open-source models for table-text QA tasks.

Table~\ref{Performance of OTT-QA and related works} shows the performance of our model - TTQA-RS on the OTT-QA development set. To the best of our knowledge, MFORT-QA \citep{guan2024mfort} is the only model that has used in-context learning for the OTT-QA dataset, but since they have not reported their performance on the development set, we therefore compare our model's performance with other existing works that trained the models. Our TTQA-RS model with LLaMA 3-70B with self-consistency has achieved new state-of-the-art performance on the OTT-QA's development set.

\subsection{Analysis and Ablation Studies}

\subsubsection{Effect of converting rows to sentences}

We assessed the retriever model's effectiveness on the HybridQA test set under two conditions: first, when the table rows are converted into sentences prior to retrieval, and second, when the table rows are used in their original form without conversion. We obtained HIT@1 of 71.49 when rows were converted to sentences and HIT@1 of 60.75 when the tables were directly fed to the table retriever. This shows that converting the rows to sentences before feeding them to the retriever improves the performance of our table-text QA retriever model.

\subsubsection{Effect of reader} 
Table~\ref{Ablation studies of TTQA-RS on HybridQA dataset} shows the ablation studies of our model using HybridQA dataset and OTT-QA's development set. We can observe that baseline CoT model outperforms the baseline standard prompting model. We tested the model by adding each component of our reader one by one, and we notice a significant increase in the performance of the model for all the datasets. This shows the importance of every component of our model. In Appendix ~\ref{Impact of number of shots}, we evaluated the impact of the number of shots on the model's performance.



\subsection{Human Evaluation Results}
We manually evaluated the first 80 samples of the table-text summaries generated by LLaMA3-70B, and 100 samples of the independent sub-questions generated using LLM prompting were also evaluated. We obtained an accuracy of 91\% for question decomposition. For evaluating the generated summaries using retrieved table rows and passages, we have used three evaluation metrics - correctness, inclusivity, and completeness. For correctness, we checked if the summary generated is overall correct and if the model generates any hallucination. For inclusivity, we checked if the generated summaries included information about both the retrieved rows and passages. Completeness was used to check if the generated summaries had complete sentences. We obtained a correctness score, inclusivity score, and completeness score of 0.83, 0.87, and 0.98, respectively, for the 80 HybridQA test set samples.

\section{Conclusion}
This paper proposes a RAG-based framework with an enhanced retriever by converting rows to sentences before feeding the tables to a retriever. We have further proposed a prompting strategy of multi-hop table-text QA by generating table-text summaries and answers of sub-questions.  We show that including summaries of retrieved table rows and passages in the prompt with our breakdown approach can substantially increase the performance of CoT prompting in table-text QA. The proposed method achieves new state-of-the-art performance among the prompting approaches for multi-hop table-text QA tasks using open-source (i.e. LLaMA3-70B) models.

\section*{Limitations}

Our work has several limitations. Firstly, we are breaking down our problem into individual steps. Even though breaking down the problem into sub-problems helps to reduce hallucination while reasoning with open-source LLMs, it also causes error propagation. Errors made in the initial steps can result in wrong answers. Secondly, the performance of our prompting-based approach, even though is on par with the fine-tuned state-of-the-art models (or has outperformed the training-based state-of-the-model for OTT-QA development set), it's performance is still not close to the supervised models (in case of HybridQA dataset) and also not close to human performance. Also, currently, we have only experimented with multi-hop table-text datasets in which the questions are already linked to the tables. The test set of the OTT-QA dataset does not have links between the tables with texts. This is out of scope of this current work, but in the future, we plan to explore more in this area.

\bibliography{paper_arxiv2}

\begin{thebibliography}{22}
\providecommand{\natexlab}[1]{#1}

\bibitem[{Brown et~al.(2020)Brown, Mann, Ryder, Subbiah, Kaplan, Dhariwal, Neelakantan, Shyam, Sastry, Askell et~al.}]{brown2020language}
Tom Brown, Benjamin Mann, Nick Ryder, Melanie Subbiah, Jared~D Kaplan, Prafulla Dhariwal, Arvind Neelakantan, Pranav Shyam, Girish Sastry, Amanda Askell, et~al. 2020.
\newblock Language models are few-shot learners.
\newblock \emph{Advances in neural information processing systems}, 33:1877--1901.

\bibitem[{Chen(2023)}]{chen2023large}
Wenhu Chen. 2023.
\newblock Large language models are few (1)-shot table reasoners.
\newblock In \emph{Findings of the Association for Computational Linguistics: EACL 2023}, pages 1120--1130.

\bibitem[{Chen et~al.(2020{\natexlab{a}})Chen, Chang, Schlinger, Wang, and Cohen}]{chen2020open}
Wenhu Chen, Ming-Wei Chang, Eva Schlinger, William Wang, and William~W Cohen. 2020{\natexlab{a}}.
\newblock Open question answering over tables and text.
\newblock \emph{arXiv preprint arXiv:2010.10439}.

\bibitem[{Chen et~al.(2020{\natexlab{b}})Chen, Zha, Chen, Xiong, Wang, and Wang}]{chen2020hybridqa}
Wenhu Chen, Hanwen Zha, Zhiyu Chen, Wenhan Xiong, Hong Wang, and William~Yang Wang. 2020{\natexlab{b}}.
\newblock Hybridqa: A dataset of multi-hop question answering over tabular and textual data.
\newblock In \emph{Findings of the Association for Computational Linguistics: EMNLP 2020}, pages 1026--1036.

\bibitem[{Eisenschlos et~al.(2021)Eisenschlos, Gor, M{\"u}ller, and Cohen}]{eisenschlos-etal-2021-mate}
Julian Eisenschlos, Maharshi Gor, Thomas M{\"u}ller, and William Cohen. 2021.
\newblock \href {https://doi.org/10.18653/v1/2021.emnlp-main.600} {{MATE}: Multi-view attention for table transformer efficiency}.
\newblock In \emph{Proceedings of the 2021 Conference on Empirical Methods in Natural Language Processing}, pages 7606--7619. Association for Computational Linguistics.

\bibitem[{Feng et~al.(2022)Feng, Han, Sun, and Li}]{feng-etal-2022-multi}
Yue Feng, Zhen Han, Mingming Sun, and Ping Li. 2022.
\newblock \href {https://doi.org/10.18653/v1/2022.findings-naacl.12} {Multi-hop open-domain question answering over structured and unstructured knowledge}.
\newblock In \emph{Findings of the Association for Computational Linguistics: NAACL 2022}, pages 151--156. Association for Computational Linguistics.

\bibitem[{Guan et~al.(2024)Guan, Huang, and Zhang}]{guan2024mfort}
Che Guan, Mengyu Huang, and Peng Zhang. 2024.
\newblock Mfort-qa: Multi-hop few-shot open rich table question answering.
\newblock \emph{arXiv preprint arXiv:2403.19116}.

\bibitem[{Herzig et~al.(2020)Herzig, Nowak, M{\"u}ller, Piccinno, and Eisenschlos}]{herzig-etal-2020-tapas}
Jonathan Herzig, Pawel~Krzysztof Nowak, Thomas M{\"u}ller, Francesco Piccinno, and Julian Eisenschlos. 2020.
\newblock \href {https://doi.org/10.18653/v1/2020.acl-main.398} {{T}a{P}as: Weakly supervised table parsing via pre-training}.
\newblock In \emph{Proceedings of the 58th Annual Meeting of the Association for Computational Linguistics}, pages 4320--4333. Association for Computational Linguistics.

\bibitem[{Kumar et~al.(2023)Kumar, Gupta, Chemmengath, Sen, Chakrabarti, Bharadwaj, and Pan}]{kumar-etal-2023-multi}
Vishwajeet Kumar, Yash Gupta, Saneem Chemmengath, Jaydeep Sen, Soumen Chakrabarti, Samarth Bharadwaj, and Feifei Pan. 2023.
\newblock \href {https://doi.org/10.18653/v1/2023.acl-long.449} {Multi-row, multi-span distant supervision for {T}able+{T}ext question answering}.
\newblock In \emph{Proceedings of the 61st Annual Meeting of the Association for Computational Linguistics (Volume 1: Long Papers)}, pages 8080--8094. Association for Computational Linguistics.

\bibitem[{Lee et~al.(2023)Lee, Park, Seo, Jeon, Kang, and Na}]{lee-etal-2023-mafid}
Sung-Min Lee, Eunhwan Park, Daeryong Seo, Donghyeon Jeon, Inho Kang, and Seung-Hoon Na. 2023.
\newblock \href {https://doi.org/10.18653/v1/2023.findings-eacl.177} {{MAF}i{D}: Moving average equipped fusion-in-decoder for question answering over tabular and textual data}.
\newblock In \emph{Findings of the Association for Computational Linguistics: EACL 2023}, pages 2337--2344. Association for Computational Linguistics.

\bibitem[{Lei et~al.(2023)Lei, Li, Wei, He, Huang, Zhao, and Liu}]{lei-etal-2023-s3hqa}
Fangyu Lei, Xiang Li, Yifan Wei, Shizhu He, Yiming Huang, Jun Zhao, and Kang Liu. 2023.
\newblock \href {https://doi.org/10.18653/v1/2023.acl-short.147} {{S}3{HQA}: A three-stage approach for multi-hop text-table hybrid question answering}.
\newblock In \emph{Proceedings of the 61st Annual Meeting of the Association for Computational Linguistics (Volume 2: Short Papers)}, pages 1731--1740. Association for Computational Linguistics.

\bibitem[{Li et~al.(2021)Li, Ng, Xu, Zhu, Wang, and Xiang}]{li-etal-2021-dual}
Alexander~Hanbo Li, Patrick Ng, Peng Xu, Henghui Zhu, Zhiguo Wang, and Bing Xiang. 2021.
\newblock \href {https://doi.org/10.18653/v1/2021.acl-long.315} {Dual reader-parser on hybrid textual and tabular evidence for open domain question answering}.
\newblock In \emph{Proceedings of the 59th Annual Meeting of the Association for Computational Linguistics and the 11th International Joint Conference on Natural Language Processing (Volume 1: Long Papers)}, pages 4078--4088. Association for Computational Linguistics.

\bibitem[{Pasupat and Liang(2015)}]{pasupat-liang-2015-compositional}
Panupong Pasupat and Percy Liang. 2015.
\newblock \href {https://doi.org/10.3115/v1/P15-1142} {Compositional semantic parsing on semi-structured tables}.
\newblock In \emph{Proceedings of the 53rd Annual Meeting of the Association for Computational Linguistics and the 7th International Joint Conference on Natural Language Processing (Volume 1: Long Papers)}, pages 1470--1480. Association for Computational Linguistics.

\bibitem[{Sun et~al.(2021)Sun, Cohen, and Salakhutdinov}]{sun2021end}
Haitian Sun, William~W Cohen, and Ruslan Salakhutdinov. 2021.
\newblock End-to-end multihop retrieval for compositional question answering over long documents.
\newblock \emph{arXiv preprint arXiv:2106.00200}.

\bibitem[{Wang et~al.(2022{\natexlab{a}})Wang, Wei, Schuurmans, Le, Chi, Narang, Chowdhery, and Zhou}]{wang2022self}
Xuezhi Wang, Jason Wei, Dale Schuurmans, Quoc Le, Ed~Chi, Sharan Narang, Aakanksha Chowdhery, and Denny Zhou. 2022{\natexlab{a}}.
\newblock Self-consistency improves chain of thought reasoning in language models.
\newblock \emph{arXiv preprint arXiv:2203.11171}.

\bibitem[{Wang et~al.(2022{\natexlab{b}})Wang, Bao, Duan, Wu, He, and Zhao}]{wang-etal-2022-muger2}
Yingyao Wang, Junwei Bao, Chaoqun Duan, Youzheng Wu, Xiaodong He, and Tiejun Zhao. 2022{\natexlab{b}}.
\newblock \href {https://doi.org/10.18653/v1/2022.findings-emnlp.498} {{M}u{GER}2: Multi-granularity evidence retrieval and reasoning for hybrid question answering}.
\newblock In \emph{Findings of the Association for Computational Linguistics: EMNLP 2022}, pages 6687--6697. Association for Computational Linguistics.

\bibitem[{Wei et~al.(2022)Wei, Wang, Schuurmans, Bosma, Xia, Chi, Le, Zhou et~al.}]{wei2022chain}
Jason Wei, Xuezhi Wang, Dale Schuurmans, Maarten Bosma, Fei Xia, Ed~Chi, Quoc~V Le, Denny Zhou, et~al. 2022.
\newblock Chain-of-thought prompting elicits reasoning in large language models.
\newblock volume~35, pages 24824--24837.

\bibitem[{Yin et~al.(2020)Yin, Neubig, Yih, and Riedel}]{yin-etal-2020-tabert}
Pengcheng Yin, Graham Neubig, Wen-tau Yih, and Sebastian Riedel. 2020.
\newblock \href {https://doi.org/10.18653/v1/2020.acl-main.745} {{T}a{BERT}: Pretraining for joint understanding of textual and tabular data}.
\newblock In \emph{Proceedings of the 58th Annual Meeting of the Association for Computational Linguistics}, pages 8413--8426. Association for Computational Linguistics.

\bibitem[{Yu et~al.(2018)Yu, Zhang, Yang, Yasunaga, Wang, Li, Ma, Li, Yao, Roman, Zhang, and Radev}]{yu-etal-2018-spider}
Tao Yu, Rui Zhang, Kai Yang, Michihiro Yasunaga, Dongxu Wang, Zifan Li, James Ma, Irene Li, Qingning Yao, Shanelle Roman, Zilin Zhang, and Dragomir Radev. 2018.
\newblock \href {https://doi.org/10.18653/v1/D18-1425} {{S}pider: A large-scale human-labeled dataset for complex and cross-domain semantic parsing and text-to-{SQL} task}.
\newblock In \emph{Proceedings of the 2018 Conference on Empirical Methods in Natural Language Processing}, pages 3911--3921, Brussels, Belgium. Association for Computational Linguistics.

\bibitem[{Zhong et~al.(2017)Zhong, Xiong, and Socher}]{zhong2017seq2sql}
Victor Zhong, Caiming Xiong, and Richard Socher. 2017.
\newblock Seq2sql: Generating structured queries from natural language using reinforcement learning.
\newblock \emph{arXiv preprint arXiv:1709.00103}.

\bibitem[{Zhong et~al.(2022)Zhong, Huang, Liu, Zhou, Wang, Yin, and Duan}]{zhong2022reasoning}
Wanjun Zhong, Junjie Huang, Qian Liu, Ming Zhou, Jiahai Wang, Jian Yin, and Nan Duan. 2022.
\newblock Reasoning over hybrid chain for table-and-text open domain qa.
\newblock In \emph{Proceedings of the Thirty-First International Joint Conference on Artificial Intelligence (IJCAI-22)}.

\bibitem[{Zhou et~al.(2022)Zhou, Sch{\"a}rli, Hou, Wei, Scales, Wang, Schuurmans, Cui, Bousquet, Le et~al.}]{zhou2022least}
Denny Zhou, Nathanael Sch{\"a}rli, Le~Hou, Jason Wei, Nathan Scales, Xuezhi Wang, Dale Schuurmans, Claire Cui, Olivier Bousquet, Quoc Le, et~al. 2022.
\newblock Least-to-most prompting enables complex reasoning in large language models.
\newblock \emph{arXiv preprint arXiv:2205.10625}.

\end{thebibliography}

\appendix


\clearpage
\appendix
\addcontentsline{toc}{section}{Appendix}
\section*{Appendix}
\section{Implementation Details}
\label{Implementation Details}

For all our experiments, we used Nvidia Geforce GTX 1660 Ti. For the retriever framework, we used a pre-trained BART-Large model for converting rows to sentences, with a maximum length set to 50, and the number of returned sequences set to 1. For the table retriever, the DPR is fine-tuned with batch size of 32 and trained for 8 epochs, with a learning rate of 2e-5. We trained the model for 5000 questions which took 6 hours to train the model. For the re-ranker model, we have used the pre-trained 'deepset/roberta-base-squad2' model with the maximum length set to 512, and number of returned sequences set to 1. We have set the $\alpha$ to 0.2 for our experiment.

For the reader, we have set the temperature to 0.5 for all experiments that use zero-shot or few-shot learning. The table-text summary was generated using zero-shot learning with LLaMA 3-70B and for the question decomposition step, we used two-shot learning with LLAMA3B-70B. This was consistent for all our experiments. In the last two stages of our TTQA-RS framework, i.e. for steps involving table-text QA of independent sub-questions and for table-text QA of original questions, we have used few-shot learning with CoT, and we have experimented with different language models such as LLaMA 2-7b, LLaMA2-13b, LLaMA2-70b, LLaMA3-8b, and LLaMA3-70b.

\section{Table-text summarization prompt}
\label{Table-text summarization}

\begin{figure}[hbt!]
    \centering
    \includegraphics[width=1\linewidth]{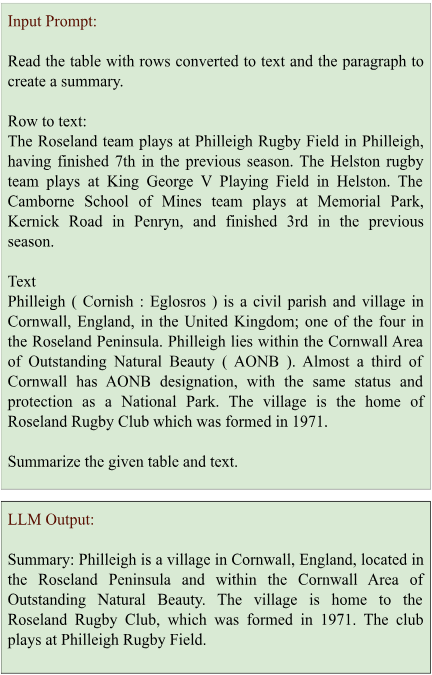}
    \caption{Prompt for zero-shot table-text summarization}
    \label{fig:summarization prompt}
\end{figure}

Figure \ref{fig:summarization prompt} shows an example of a table-text summarization prompt. The LLM output shows the generated summary.

\section{Independent sub-questions}
\label{Independent sub-questions}
Table ~\ref{tab:Examples of complex questions and independent sub-questions} shows examples of original complex questions and their corresponding independent sub-questions.  The sub-answer obtained using the independent sub-question aids in answering the original complex question.

\setlength{\arrayrulewidth}{0.5mm}
\setlength{\tabcolsep}{6pt}
\renewcommand{\arraystretch}{1.1}
\begin{table*}[hbt!]
    \centering
  \caption{Examples of complex questions and independent sub-questions}
  \label{tab:Examples of complex questions and independent sub-questions}
  \begin{tabular}{|M{7.2cm}|M{7cm}|}
    \hline
    \textbf{Original Complex Question} & \textbf{Independent sub-question}  \\
     \hline
   What was the release date of the game which Andrew Voss provided commentary on ?  & What game has Andrew Voss provided commentary on ? \\
    \hline
    What is the population of the hometown of the 2012 Gatorade Player of the Year ? & What is hometown of the 2012 Gatorade Player of the Year ? \\
    \hline
    Which battle took place in largest country in sub-Saharan Africa ?  & What is the largest country in sub-Saharan Africa ? \\
    \hline
    What did the Japanese rename the road that Hop Yat Church is located along ? &  Where is Hop Yat Church located? \\
    \hline
    What is the genre of the game that began development as Street Fighter II , but had its name and genre changed ? &  Which game began development as Street Fighter II, but had its name and genre changed? \\
    \hline
    What was the highest legal job of the chancellor of Wilfrid Laurier University after May 1990 ? & Who was the chancellor of Wilfrid Laurier University after May 1990? \\
    \hline
    What is the availability status of the Buffalo station owned by the E. W. Scripps Company ? & Which station in Buffalo is owned by the E. W. Scripps Company? \\
    \hline
    Which war film was directed by the director of Hibiscus Town ? & Who directed Hibiscus Town? \\
    \hline
    How many plays has the person known for The Black Coat written ? & Who is the person known for The Black Coat? \\

    \hline
  \end{tabular}

\end{table*}

\section{Impact of number of shots}
\label{Impact of number of shots}
\begin{figure*}
    \centering
    \includegraphics[scale=0.60]{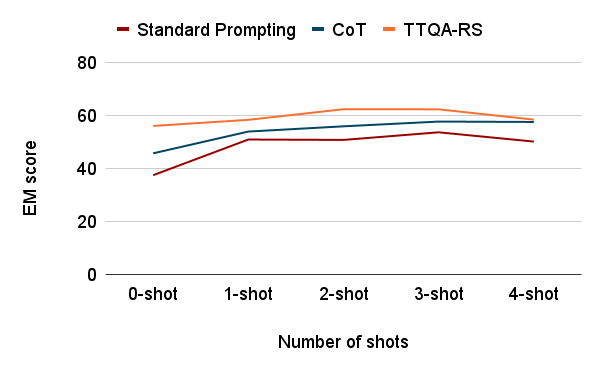}
    \caption{k-shot ablation study over Hybrid-QA test set}
    \label{fig:k-shot ablation study over Hybrid-QA test set}
\end{figure*}

In this section, we have performed an ablation study by increasing the number of shots while evaluating our model on the test set of the HybridQA dataset. This is shown in Figure~\ref{fig:k-shot ablation study over Hybrid-QA test set}. We have evaluated the impact of increasing k in k-shot learning on the baseline standard prompting model, baseline CoT model, and and the TTQA-RS model using LLaMA3 -70B. For standard prompting and CoT, we observe that with an increase in k from 0 to 3, there is an increase in the exact match score. After 3 shots, increasing the number of shots does not improve the performance. For the TTQA-RS model, there is an improvement in EM score from 0-shot to 2-shot, after which increasing the k value does not improve the exact match score of the model.

\end{document}